# Optimising Random Forest Machine Learning Algorithms for User VR Experience Prediction Based on Iterative Local Search-Sparrow Search Algorithm

Xirui Tang[1]*, Feiyang Li[2], Zinan Cao[3], Qixuan Yu[4], Yulu Gong[5],

[1]College of Computer Sciences, Northeastern University, Boston, MA, 02115, USA

[2]Department of Computer Science, University of Illinois Urbana-Champaign, Champaign, IL, 61820, USA

[3]Department of General Systems Studies, The University of Tokyo, Tokyo, 113-8654, Japan

[4]College of Computing, Georgia Institute of Technology, Atlanta, GA, 30332-0280, USA

[5]School of Informatics, Computing, and Cyber Systems, Northern Arizona University, Flagstaff, AZ, 86011-4092, USA

* Corresponding author: e-mail: tang.xir@northeastern.edu

*Abstract*—In this paper, an improved method for VR user experience prediction is investigated by introducing a sparrow search algorithm and a random forest algorithm improved by an iterative local search-optimised sparrow search algorithm. The study firstly conducted a statistical analysis of the data, and then trained and tested using the traditional random forest model, the random forest model improved by the sparrow search algorithm, and the random forest algorithm improved based on the iterative local search-sparrow search algorithm, respectively. The results show that the traditional random forest model has a prediction accuracy of 93% on the training set but only 73.3% on the test set, which is poor in generalisation; whereas the model improved by the sparrow search algorithm has a prediction accuracy of 94% on the test set, which is improved compared with the traditional model. What is more noteworthy is that the improved model based on the iterative local search-sparrow search algorithm achieves 100% accuracy on both the training and test sets, which is significantly better than the other two methods. These research results provide new ideas and methods for VR user experience prediction, especially the improved model based on the iterative local search-sparrow search algorithm performs well and is able to more accurately predict and classify the user's VR experience. In the future, the application of this method in other fields can be further explored, and its effectiveness can be verified through real cases to promote the development of AI technology in the field of user experience.

*Keywords-VR; Search-sparrow search; Random forest; Machine learning; AI; Random forest;*

## I. Introduction

The user's virtual reality (VR) experience is critical to the development of VR technology. As VR technology continues to evolve, research on user experience becomes increasingly important. Understanding how users feel and react in VR environments can help improve VR applications and devices and increase user satisfaction and engagement.

Machine learning algorithms play a key role in the context of studying users' VR effect experience. By collecting a large amount of user data, machine learning algorithms can analyse the user's behavioural, emotional and physiological responses in VR environments in order to predict the user's experience [1]. These data can include the user's eye track, physiological indicators (e.g., heart rate, electrical skin activity, etc.), behavioural data (e.g., movement paths, interaction behaviours, etc.), and the user's subjective evaluation. Machine learning algorithms play a key role in predicting user VR experiences. They are able to predict the user's experience in a given context by analysing large amounts of data to discover hidden patterns and associations. For example, by monitoring a user's eye track and physiological indicators, machine learning algorithms can infer the user's attention and emotional state, and thus predict the user's level of satisfaction with the VR experience [2].

Among the many machine learning algorithms, the random forest algorithm has attracted much attention due to its advantages in handling large-scale datasets and high-dimensional feature spaces. Random forest is an integrated learning method that makes predictions by constructing multiple decision trees and combining them [3]. It has the advantages of high accuracy and resistance to overfitting, etc. Improvements based on the Random Forest algorithm can further improve the prediction of user VR experience. By optimising the parameters of the algorithm, improving the feature engineering methods or introducing new data features, the model can be made to predict the user's experience in the VR environment more accurately. This is important for optimising the design of VR applications and devices and improving the quality of user experience. In this paper, we improve the random forest algorithm based on two methods, the sparrow search algorithm and the sparrow search algorithm with iterative local search optimisation, and compare the advantages and disadvantages of each model in the prediction of the accuracy of God, which provides a variety of new methods for VR user experience prediction.

## II. Related work

With the rapid development of Virtual Reality (VR) technology, there is a growing interest in how to evaluate and enhance the VR experience. One of the important research directions is to predict the user's immersion level in VR environments through machine learning algorithms. The level of immersion is a measure of the user's engagement and perceived quality of the virtual environment, and is crucial for evaluating and improving the user experience of VR applications.

Machine learning algorithms play a key role in this area of research. Researchers typically use a variety of sensor technologies (e.g., eye tracking, heart rate monitoring, galvanic skin response, etc.) to obtain physiological and behavioural data from users in VR [4,5]. These data can include the user's gaze pattern, eye movements, heart rate variability, muscle tension, etc., thus reflecting the user's response and immersion level in the virtual environment. These data are then used as features to construct machine learning models to predict the user's level of immersion in VR, in combination with the user's subjective evaluation of the VR experience.

Researchers also often explore the performance of different types of machine learning algorithms in VR immersion prediction. For example, algorithms such as Support Vector Machines (SVMs) [6], Neural Networks (NNs) [7], and Random Forests (RFs) [8] have been applied to this task, and some studies try to combine them to obtain better prediction performance. In addition, some studies have considered the generalisation ability and real-time nature of the models in order to apply them to different types and sizes of VR applications.

In addition to the machine learning algorithms themselves, researchers are also working on developing new feature extraction methods and data preprocessing techniques to optimise the performance of the models. This may involve techniques such as feature selection, dimensionality reduction, data augmentation, etc., to ensure that the model is able to extract the most informative features from complex physiological and behavioural data, thereby improving the accuracy and robustness of predictions.

III. DATA SOURCES AND STATISTICAL ANALYSES OF DATA

The data used in this paper comes from the Kaggle open source dataset at:https: //www.kaggle.com/datasets/ aakashjoshi123/virtual-reality-experiences/data, which consists of users' virtual reality (VR) environment experience. It includes data related to physiological responses such as heart rate and skin conductance, emotional states and user preferences. The final variable to be predicted was Immersion, which was categorised into two categories, with 1 indicating an inability to immerse while 2 indicating that the user was well immersed. Some of the data is shown in Table 1.

TABLE I. SELECTED DATA SETS

| Age | Gender | VRHeadset | Duration | Motion Sickness | Immersion Level |
|---|---|---|---|---|---|
| 40 | Male | HTC Vive | 13.59 | 8 | 2 |
| 43 | Female | HTC Vive | 19.95 | 2 | 2 |
| 27 | Male | PlayStation VR | 16.54 | 4 | 2 |
| 33 | Male | HTC Vive | 42.57 | 6 | 2 |
| 51 | Male | PlayStation VR | 22.45 | 4 | 2 |
| 46 | Other | Oculus Rift | 28.28 | 7 | 2 |
| 49 | Other | Oculus Rift | 52.29 | 7 | 2 |
| 42 | Other | Oculus Rift | 22.95 | 8 | 2 |
| 56 | Female | PlayStation VR | 16.85 | 5 | 1 |
| 60 | Male | Oculus Rift | 34.47 | 5 | 2 |
| 37 | Male | Oculus Rift | 43.36 | 2 | 1 |

The data were first statistically analysed for the maximum, minimum, mean, standard deviation, median and variance of each variable and the results are shown in Table 2.

TABLE II. STATISTICAL RESULTS

| Variable Name | Maximum | Minimum | Mean | Standard |
|---|---|---|---|---|
| Age | 60 | 18 | 39.178 | 12.05 | 39 |
| Gender | 3 | 1 | 2.017 | 0.822 | 2 |
| VR Headset | 3 | 1 | 2.009 | 0.823 | 2 |
| Duration | 59 | 5 | 32.577 | 15.765 | 32.369 |
| Motion Sickness | 10 | 1 | 5.526 | 2.867 | 6 |

After counting the data, it was found that there were no vacancies, then we firstly eliminated the data with duplicate values and used the 3σ principle to determine the outliers and eliminated the data with outliers.

IV. METHOD

A. Random forest

Random Forest is an integrated learning method for prediction or classification by constructing multiple decision trees. Its principle is based on the Bagging algorithm (Bootstrap Aggregating) and random feature selection. Specifically, Random Forest consists of multiple decision trees, each trained on a different training set. When constructing each tree, a certain number of samples are randomly drawn from the original training set with putback as the training set for that tree, which can increase the diversity of the model [9]. The model structure diagram of random forest is shown in Figure 1.

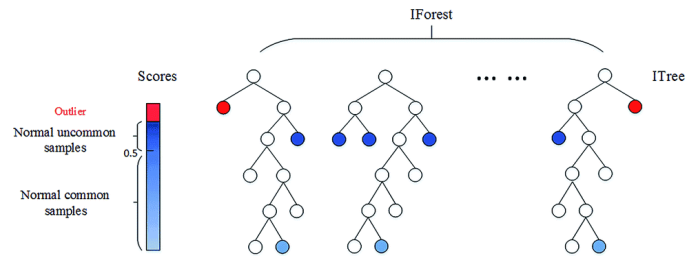

Figure 1. The model structure diagram of random forest.

In the process of constructing each decision tree, Random Forest also performs random selection of features. That is, when splitting nodes each time, only a random subset of features are considered to find the best splitting point, which helps to reduce the variance of the model and improve the generalisation ability. Eventually, when a prediction is needed, each decision tree in the random forest will give its own prediction, and then the final prediction will be obtained by voting or averaging.

B. Sparrow Search Algorithm

The Sparrow Search Algorithm (SSA) is an emerging heuristic optimisation algorithm inspired by the behaviour of sparrows when searching for food. The algorithm simulates the strategy of sparrows when searching for food, and achieves the search of the global optimal solution through information sharing and learning among individuals. In the algorithm, each individual represents a candidate solution and iteratively updates itself according to its own state and neighbourhood information, thus gradually approaching the optimal solution. The schematic diagram of the sparrow search algorithm is shown in Fig. 2.

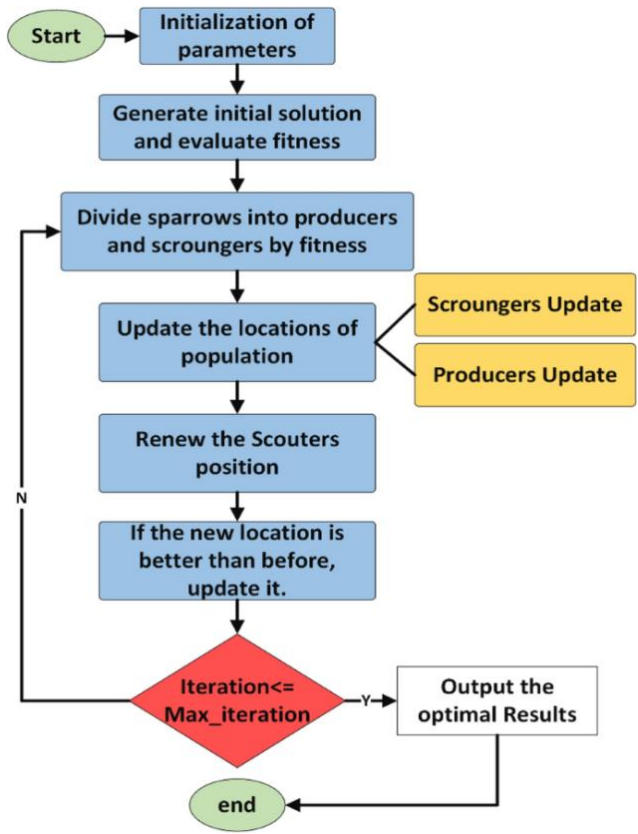

Figure 2.  The schematic diagram of the sparrow search algorithm.

The main principle of the sparrow search algorithm includes three key steps: foraging, communication and learning. In the foraging phase, the individual determines the next moving direction according to its own position and fitness value; in the communication phase, the individual accelerates the search process by exchanging information; and in the learning phase, the individual adjusts its own state according to the information of its neighbours and its historical experience. Through these three stages, the whole population gradually converges to the neighbourhood of the optimal solution.

## C. Improved sparrow search algorithm based on iterative local search

Improved sparrow search algorithm based on iterative local search is a heuristic optimisation algorithm mainly used to solve combinatorial optimisation problems. Firstly, an initial solution is randomly generated as the current optimal solution and an upper limit is set on the number of iterations. Then, in each round of iterations, a local search strategy is used to find a better solution in the neighbourhood of the current optimal solution and update the current optimal solution. Next, the choice of whether to accept the new solution is made according to a certain probability to avoid falling into the local optimal solution. The above steps are repeated until the upper limit of the number of iterations is reached or the stopping condition is satisfied [10]. The algorithm schematic of the improved sparrow search algorithm based on iterative local search is shown in Fig. 3.

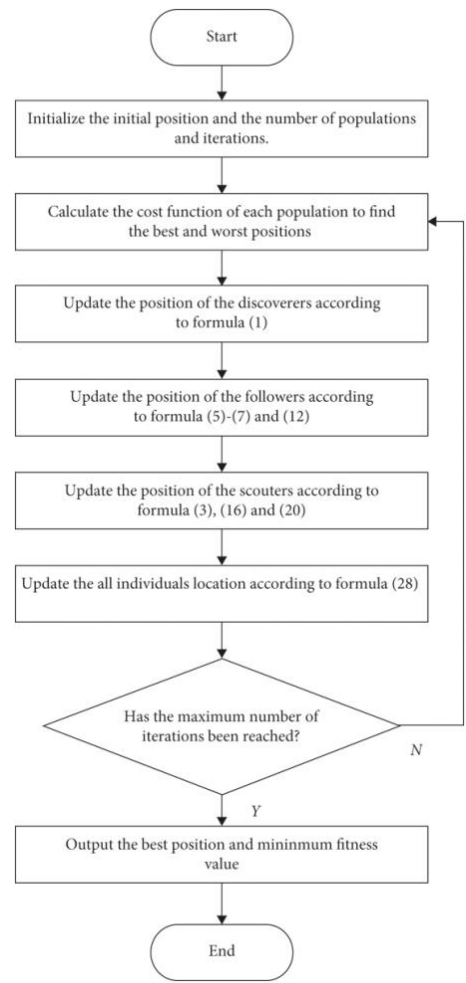

Figure 3.  The algorithm schematic of the improved sparrow search algorithm based on iterative local search.

The improved sparrow search algorithm introduces a more effective strategy in the local search process compared to the traditional sparrow search algorithm, adopting more appropriate neighbourhood search methods, adjusting the parameter settings, and increasing the convergence control in order to improve the algorithm's convergence speed and global search capability. By constantly iterating and updating the current optimal solution, the algorithm can gradually approach or even reach the global optimal solution, so as to achieve better optimisation results in practical problems.

## V. RESULT

For the experimental part, Matlab R2023a was chosen for the experiments in this project, which were trained using the random forest model, the random forest model improved by the sparrow search algorithm, and the random forest algorithm improved by the iterative local search-sparrow search algorithm, respectively. For the division of the dataset, all three experiments were divided using a 7:3 ratio, and the learning rate was set to 0.001.

The sparrow search algorithm, when applied to the random forest model, uses its population intelligence and local search properties to optimise aspects of the decision tree. Firstly, the sparrow search algorithm is used to adjust the parameter settings, i.e. tree depth and node splitting criterion, etc., during the decision tree building process to improve the performance of the model. Secondly, the algorithm is utilised in the feature selection phase for searching and selecting a subset of features so as to improve the model's ability to adapt to the data features. There are three improvements in the sparrow search algorithm for local search optimisation, which are population initialisation using chaotic mapping, dynamic adaptive weighting, and improvement of the optimal sparrow iteration using backward learning and Cauchy variation.

In the determination of the number of iterations, at the beginning of this paper, the epoch of the three models is uniformly set to 100, and observe the change of the loss, and it is found that the three models begin to converge in the 20th to the 25th epoch or so, in order to avoid the occurrence of the overfitting phenomenon, this paper will set the epoch to 40 to reexperiment on the three models.

After the model training, the prediction confusion matrices of the training and test sets of the three algorithms are outputted respectively, the prediction confusion matrices of the training and test sets of the Random Forest model are shown in Fig. 4, the prediction confusion matrices of the training and test sets of the Random Forest model improved by the Sparrow search algorithm are shown in Fig. 5, and the prediction confusion matrices of the training and test sets of the Random Forest model improved based on the iterative local search - Sparrow search algorithm are shown in Fig. 6. matrix is shown in Fig. 6, and the prediction accuracies of the test sets of the three models are shown in Table 3.

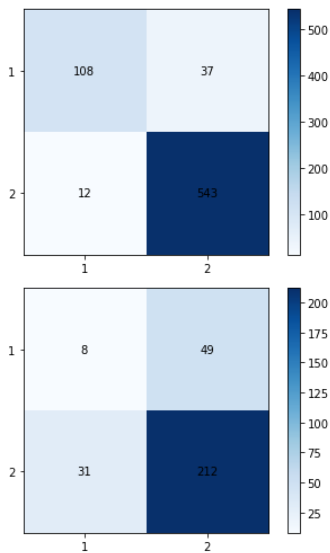

Figure 4. Confusion matrices of the training and test sets of the Random Forest.

The prediction results of the random forest model show that the prediction accuracy of the training set is 93% and the prediction accuracy of the test set is 73.3%, and the accuracy of the test set decreases by 19.7% relative to that of the test set, which indicates that the random forest model has poor generalisation ability for VR experience prediction.

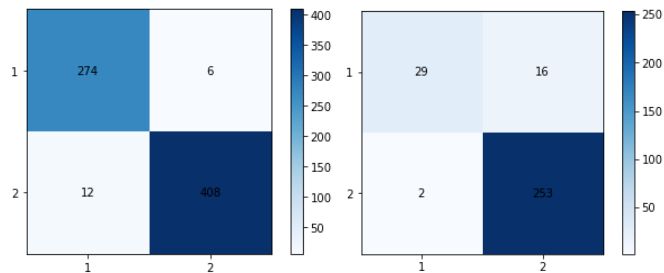

Figure 5. Confusion matrices of the training and test sets of the Random Forest model improved by the Sparrow search algorithm.

The prediction results of the random forest model improved by the sparrow search algorithm show that the prediction accuracy of the training set is 97.5% and the prediction accuracy of the test set is 94%, which is better compared to the random forest model for both the training and test sets.

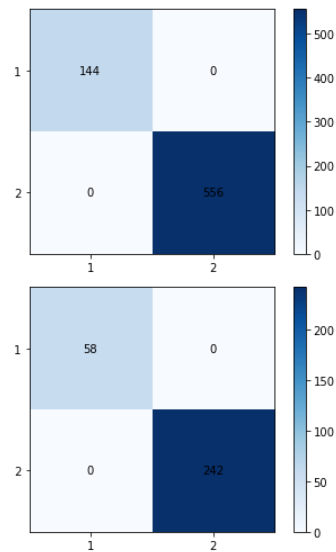

Figure 6. Confusion matrices of the training and test sets of the Random Forest model improved based on the iterative local search - Sparrow search algorithm .

The prediction results of the random forest algorithm improved based on iterative local search-sparrow search algorithm show that the accuracy of both the training set and the test set reaches 100%, which is able to predict and classify the user's VRs well, and the accuracy is significantly higher than that of the random forest model and the random forest model improved by the sparrow search algorithm.

TABLE III. THE PREDICTION ACCURACIES OF THE TEST SETS OF THE THREE MODELS

|  | RF | SSA-RF | ISSA-RF |
|---|---|---|---|
| Train | 93% | 97.5% | 100% |
| Test | 73.3% | 94% | 100% |

## VI. Conclusion

In this study, the random forest algorithm was improved by introducing the sparrow search algorithm and the iterative local search-optimised sparrow search algorithm, and explored in the field of VR user experience prediction. Firstly, the data were statistically analysed, and then the traditional random forest model, the random forest model improved by the sparrow search algorithm, and the random forest algorithm improved based on the iterative local search-sparrow search algorithm were used for training and testing, respectively.

The traditional random forest model performs well on the training set with a prediction accuracy of 93%, but performs only 73.3% on the test set, a decrease of 19.7% relative to the training set. This indicates that the traditional random forest model has insufficient generalisation ability in VR experience prediction.

The Random Forest model improved by the Sparrow Search algorithm performs well on the test set with a prediction accuracy of 94%, which is an improvement over the traditional model. It also outperforms the traditional random forest model on the training set with an accuracy of 97.5%.

The improved Random Forest algorithm based on the iterative local search-sparrow search algorithm, on the other hand, shows the most excellent performance, with an accuracy of 100% on both the training and test sets. This indicates that the method is able to predict and classify user VR experiences more accurately, significantly outperforming both the traditional random forest model and the improved model based on the simple sparrow search algorithm.

Therefore, the random forest algorithm improved based on iterative local search-sparrow search algorithm proposed in this study provides a novel and efficient method for VR user experience prediction. The method not only improves the prediction accuracy, but also enhances the generalisation ability of the model, which provides strong support for improving the quality of VR application experience. In the future, we can further explore the application potential of this method in other fields and validate it with practical scenarios, in order to promote the development and application of AI technology in the field of user experience.

## References

[1] Anwar, Muhammad Shahid, et al. "Subjective QoE of 360-degree virtual reality videos and machine learning predictions." IEEE Access 8 (2020): 148084-148099.

[2] David-John, Brendan, et al. "Towards gaze-based prediction of the intent to interact in virtual reality." ACM Symposium on Eye Tracking Research and Applications. 2021.

[3] Schwind, Valentin, et al. "The effects of full-body avatar movement predictions in virtual reality using neural networks." Proceedings of the 26th ACM Symposium on Virtual Reality Software and Technology. 2020.

[4] Hobson, J. Allan, Charles C-H. Hong, and Karl J. Friston. "Virtual reality and consciousness inference in dreaming." Frontiers in psychology 5 (2014): 113003.

[5] Gamage, Nisal Menuka, et al. "So predictable! continuous 3d hand trajectory prediction in virtual reality." The 34th Annual ACM Symposium on User Interface Software and Technology. 2021.

[6] Tasnim, Umama, et al. "Investigating Personalization Techniques for Improved Cybersickness Prediction in Virtual Reality Environments." (2024).

[7] Prawiro, Herman, et al. "Towards Efficient Visual Attention Prediction for 360 Degree Videos." 2024 IEEE International Conference on Artificial Intelligence and eXtended and Virtual Reality (AIxVR). IEEE, 2024.

[8] Ritchie, Aaron, and Harvey Ho. "Virtual reality-based meat cut planning for lamb carcasses." New Zealand Journal of Agricultural Research (2024): 1-7.

[9] Mao, Makara, Hongly Va, and Min Hong. "Video Classification of Cloth Simulations: Deep Learning and Position-Based Dynamics for Stiffness Prediction." Sensors 24.2 (2024): 549.

[10] Vitucci, Carlo. Ambient Temperature Prediction for Embedded Systems using Machine Learning. Diss. Mälardalens universitet, 2024.